\documentclass[acmtog,nonacm]{acmart}
\usepackage{booktabs} 
\usepackage{bm}
\usepackage{multirow}
\usepackage{pifont}
\usepackage{hyperref}
\usepackage{cleveref}

\usepackage[ruled]{algorithm2e} 

\SetAlFnt{\small}
\SetAlCapFnt{\small}
\SetAlCapNameFnt{\small}
\SetAlCapHSkip{0pt}

\acmJournal{TOG}

\begin{document}
\title{MIND: Multi-Scale Intent Diffusion for Text-Driven Physics-Based Humanoid Control}

\author{Bin Li}
\authornote{Equal contribution.}
\affiliation{%
 \institution{ShanghaiTech University}
 \country{China}}
\email{libin3@shanghaitech.edu.cn}
\author{Ruichi Zhang}
\authornotemark[1]
\affiliation{%
 \institution{University of Pennsylvania}
 \country{USA}}
\email{rczhang@seas.upenn.edu}
\author{Han Liang}
\authornote{Corresponding author.}
\affiliation{%
 \institution{Bytedance Seed}
 \country{China}}
\email{lianghan@shanghaitech.edu.cn}
\author{Jingyan Zhang}
\affiliation{%
 \institution{ShanghaiTech University}
 \country{China}}
\email{zhangjy7@shanghaitech.edu.cn}
\author{Juze Zhang}
\affiliation{%
 \institution{Stanford University}
 \country{USA}}
\email{juze@stanford.edu}
\author{Xin Chen}
\affiliation{%
 \institution{Bytedance Seed}
 \country{USA}}
\email{chenxin2@shanghaitech.edu.cn}
\author{Jingya Wang}
\authornotemark[2]
\affiliation{%
 \institution{ShanghaiTech University, InstAdapt}
 \country{China}
}
\email{wangjingya@shanghaitech.edu.cn}

\begin{abstract}
    Enabling physics-based humanoids to execute diverse behaviors from high-level textual commands remains a significant challenge. 
    Existing methods typically follow either a two-stage paradigm that combines kinematic motion generation with physics-based tracking, or an end-to-end imitation-learning paradigm that directly generates actions from text.
    However, the former suffers from the inherent domain shift between kinematic generation and physics-based tracking, while the latter struggles with the substantial modality gap between textual commands and low-level actions, limiting effective semantic alignment. Notably, humanoid states encode rich motion dynamics that are more semantically aligned with textual descriptions than low-level actions, making them a natural basis for deriving behavioral intent. Building upon this insight, we propose MIND, a novel end-to-end diffusion framework for text-driven physics-based humanoid control that leverages behavioral intent as a semantic bridge between textual commands and low-level actions. At its core, MIND introduces a multi-scale intent diffusion mechanism, where a holistic intent predictor captures global behavioral dynamics to guide overall behavior synthesis, while an immediate intent predictor provides step-wise, fine-grained signals for local behavior refinement at each diffusion step. This hierarchical intent formulation imposes a structured inductive bias for humanoid control, improving semantic alignment and behavioral naturalness. Furthermore, MIND encodes humanoid states into a latent space to enable more effective semantic intent modeling. Extensive experiments demonstrate that MIND outperforms existing methods and synthesizes coherent, physically plausible, and semantically aligned humanoid behaviors from text commands. Our code will be released to facilitate future research. Project page: \texttt{https://binlee26.github.io/MIND\_page}.
    
\end{abstract}

%
%


\begin{teaserfigure}
  \includegraphics[width=\textwidth]{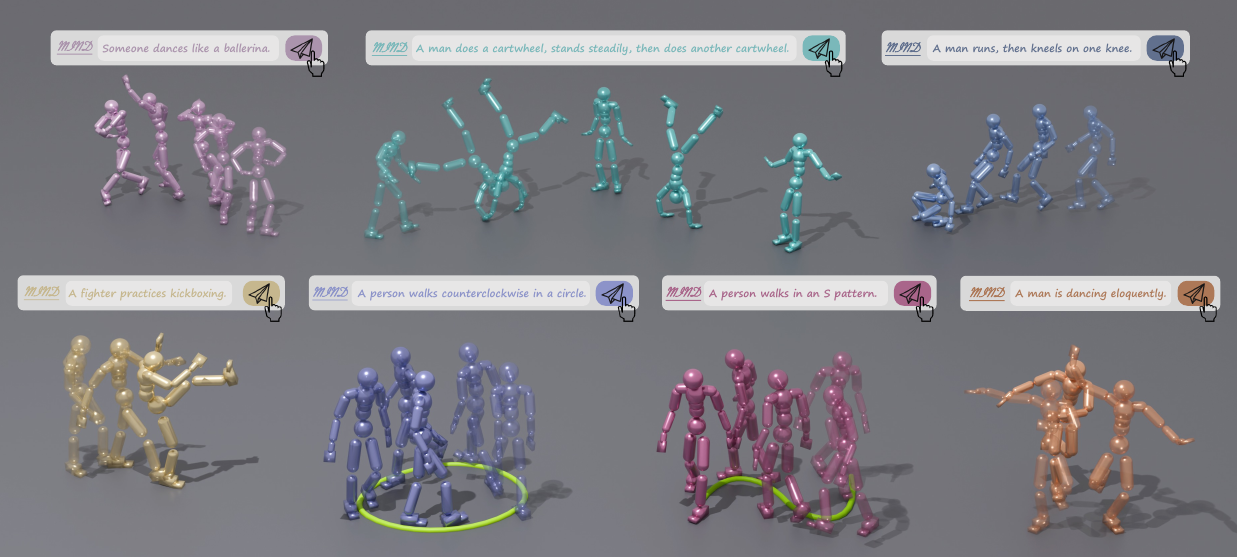}
  \caption{\textbf{MIND} is a novel end-to-end diffusion framework tailored for text-driven physics-based humanoid control, capable of   generating high-quality, diverse, and physically plausible humanoid behaviors such as dancing, kickboxing, and cartwheels.}
  \label{fig:teaser}
\end{teaserfigure}

\maketitle
\section{Introduction}
Generating controllable, natural, and physically plausible humanoid behaviors is a long-standing goal in character animation, games, embodied simulation, and robotics. 
Early works~\cite{peng2018deepmimic,peng2021amp,peng2022ase,luo2023perpetual,luo2021dynamics,luo2023universal} primarily adopt a tracking-based paradigm, where goal-conditioned reinforcement learning (RL)~\cite{liu2022goal} policies are trained to imitate reference motion trajectories. While effective, these approaches require high-quality kinematic motion data as control signals, whose acquisition is usually expensive and labor-intensive. In contrast, textual commands provide a more intuitive, flexible, and user-friendly interface for specifying humanoid behaviors, without requiring motion references or any additional predefined control signals~\cite{tessler2024maskedmimic}. However, due to the substantial modality gap between textual commands and humanoid behaviors, achieving semantically aligned and physically realistic behavioral control, as shown in Fig.~\ref{fig:teaser}, remains a non-trivial task. 

Recent works~\cite{tevet2024closd,yuan2023physdiff,xie2026textop,bao2026phygile,li2025language} attempt to handle this problem through a generation-tracking paradigm, which combines a kinematic motion planner with an RL-based tracking policy. Although these approaches achieve competitive performance, the inherent domain shift between kinematic planners and physics-based trackers makes it difficult for the tracking policy to faithfully reproduce the planned motions. While task-specific fine-tuning~\cite{tevet2024closd} has been shown to partially alleviate this discrepancy, it cannot be fully eliminated. To address this limitation, another line of work~\cite{wu2025uniphys,truong2024pdp,liao2025beyondmimic,wang2025sentinel} employs diffusion-based imitation policies~\cite{chi2025diffusion} to directly generate actions conditioned on textual commands in an end-to-end manner, thereby circumventing the issue of domain shift. Nevertheless, these existing end-to-end approaches still suffer from suboptimal text-action alignment, largely due to the absence of an effective intermediate representation for bridging the substantial modality gap between text commands and low-level motor actions. 

Notably, compared with low-level motor actions, textual semantics are more naturally manifested in humanoid states. Specifically, humanoid states inherently encode rich dynamic information of body movements, including joint configurations, motion tendencies, and temporal evolution, which collectively provide a concrete characterization of the underlying behavioral semantics. In contrast, actions mainly function as low-level motor control signals for physical actuation, with limited capacity to convey high-level behavioral semantics. Based on this, we argue that humanoid states can serve as a semantic basis for behavior modeling, where the encoded dynamic information acts as a bridge between textual commands and actions, mitigating the substantial modality gap. In this work, we refer to the dynamic information embedded in humanoid states that corresponds to underlying behavioral patterns as intent.

Building upon this insight, we propose MIND, a novel end-to-end framework for text-driven physics-based humanoid control that explicitly predicts intent as an effective semantic guidance signal. The core of MIND is a multi-scale intent mechanism that models humanoid intent at different temporal granularities within a unified diffusion framework. Specifically, a holistic intent predictor (HIP) is designed to capture overall behavioral dynamics for global planning, while an immediate intent predictor (IIP) models step-wise intent to provide fine-grained guidance for action refinement. By seamlessly integrating this hierarchical intent structure into the action diffusion transformer (ADiT), MIND effectively combines holistic planning with immediate refinement, substantially improving semantic alignment and behavioral naturalness. Furthermore, MIND encodes humanoid states into a compact latent space to obtain more structured intent representations, thereby enhancing the semantic expressiveness of the predicted intents and further improving text-action alignment. Extensive experiments demonstrate that MIND consistently outperforms several strong baselines, achieving semantically aligned and physically plausible humanoid behavior synthesis.

To summarize, our contributions are threefold:

\begin{itemize}

\item We propose MIND, a novel end-to-end framework for text-driven physics-based humanoid control, which explicitly models intent as effective semantic guidance for humanoid behavior synthesis within a unified diffusion framework.

\item We introduce a multi-scale intent mechanism that operates in a latent space and captures intent at multiple temporal granularities, enabling structured modeling of semantic dynamics.

\item Extensive experiments demonstrate the superiority of MIND over competitive baselines, and comprehensive ablation studies validate the effectiveness of each individual design.

\end{itemize}

\begin{figure}
    \centering
    \includegraphics[width=\columnwidth]{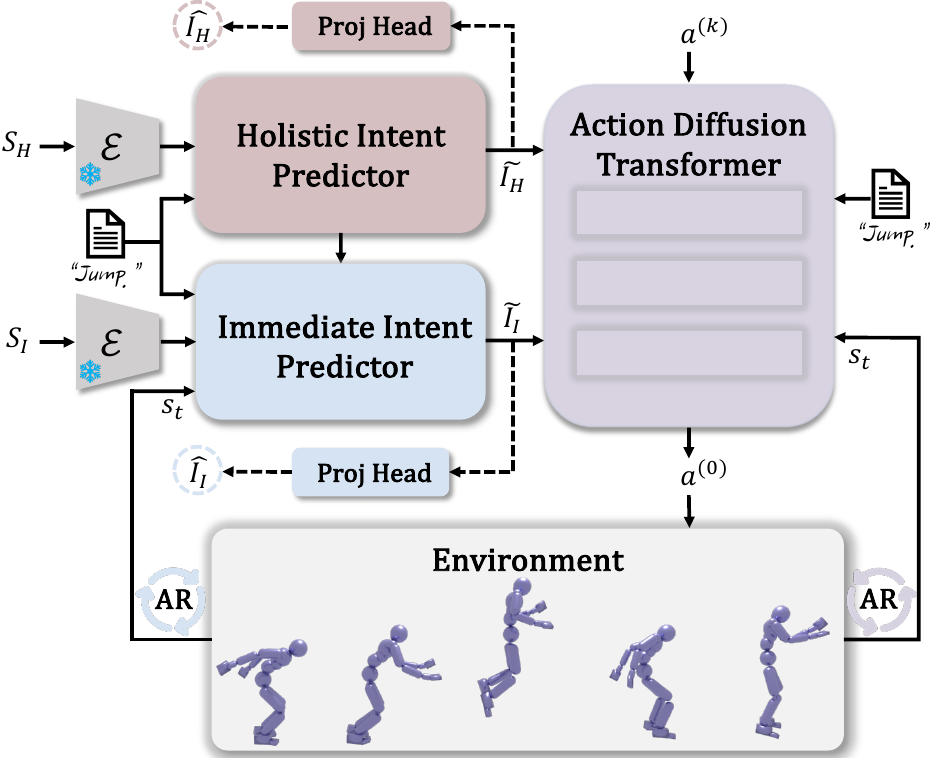}
    
    \caption{
    \textbf{Framework overview.} 
 Given text commands and humanoid states, MIND models humanoid intent at multiple temporal scales under a within diffusion framework. Specifically, the Holistic Intent Predictor (HIP) captures global behavioral dynamics to provide high-level planning guidance, while the Immediate Intent Predictor (IIP) models step-wise intent from current states for fine-grained action refinement. The predicted holistic and immediate intents, together with the textual commands, are jointly used to condition the Action Diffusion Transformer (ADiT), guiding autoregressive (AR) action generation. Dashed lines indicate components that are used only during training.
    }
    \label{fig:overview_pipeline}
\end{figure}

\section{Related Work}
\paragraph{Physics-Based Humanoid Control}
Early works on physics-based humanoid control~\cite{peng2018deepmimic} primarily focused on tracking single motion trajectories. Subsequent approaches~\cite{peng2021amp,peng2022ase,dou2023c,tessler2023calm,serifi2024vmp} extended to limited-scale motion datasets, leveraging adversarial imitation learning to produce diverse and natural behaviors. Later, methods such as UHC~\cite{luo2021dynamics}, PHC~\cite{luo2023perpetual}, and PULSE~\cite{luo2023universal} achieved impressive performance on the large-scale AMASS~\cite{mahmood2019amass} dataset. MaskedMimic~\cite{tessler2024maskedmimic} learns a versatile physics-based controller through motion inpainting. However, these approaches rely on reference motion trajectories or predefined control signals, which are often costly to acquire in practice. In contrast, textual commands provide a more accessible and user-friendly interface for behavior specification. More recent methods~\cite{huang2025diffuse,xu2025parc,wang2025hil,zhang2025physics,mu2025smp} have demonstrated promising progress, but still struggle to achieve robust text-driven control.
\paragraph{Text-Driven Behavior Synthesis}

In recent years, a large body of work has explored text-conditioned behavior synthesis for humanoid control. A dominant line of approaches focuses on kinematic motion modeling~\cite{jiang2023motiongpt,zhu2025motiongpt3,chen2025language,chen2023executing,he2025molingo,yu2026causal,zhang2024motiondiffuse,tevet2022human,zhang2023remodiffuse,guo2025snapmogen,li2026llamo}. MotionGPT~\cite{jiang2023motiongpt} discretizes motion sequences into token indices via VQVAE~\cite{van2017neural}, enabling motion generation with large language models. Molingo~\cite{he2025molingo} and CMDM~\cite{yu2026causal} improve text-motion alignment by learning a semantically structured latent space. Despite their effectiveness in text-driven kinematic motion generation, these methods largely ignore physical constraints and dynamic consistency, often resulting in physically implausible artifacts such as penetration and floating. For physics-based behavior control, existing methods can be broadly categorized into two paradigms. One is the kinematic-tracking pipeline~\cite{tevet2024closd,yuan2023physdiff}, which inevitably suffers from domain shift between kinematic generation and physics-based tracking control. The other is based on imitation learning, including PADL~\cite{juravsky2022padl}, SuperPADL~\cite{juravsky2024superpadl}, PDP~\cite{truong2024pdp}, UniPhys~\cite{wu2025uniphys}, Beyondmimic~\cite{liao2025beyondmimic}, and InterAgent~\cite{li2025interagent}, which learn from expert knowledge conditioned on text. While these approaches can mitigate the domain shift issue, they introduce a new bottleneck in bridging the substantial modality gap between high-level textual commands and low-level motor actions, hindering effective control.

\paragraph{Diffusion Policies for Humanoid Control} 

Recent advances in diffusion models~\cite{ho2022classifier,peebles2023scalable,song2020denoising,lukoianov2024score,nichol2021improved} have established them as a powerful paradigm for control~\cite{wolf2025diffusion,li2024stabilizing}, owing to their powerful generative capability, effective modeling of multi-modal distributions, and flexible conditioning mechanisms. Diffusion Policies~\cite{chi2025diffusion,zhang2025flowpolicy,wang2024one} demonstrate the effectiveness of diffusion models in manipulation tasks, achieving strong performance in learning complex action distributions from demonstrations. Building upon this line of research, PDP~\cite{truong2024pdp} extends diffusion policies to physics-based humanoid control, validating its effectiveness in motion tracking and perturbation recovery. More recently, several works~\cite{wu2025uniphys, huang2025diffuse,li2025interagent,liao2025beyondmimic} have explored the joint modeling of states and actions within a single generative framework, enabling flexible task-specific control and facilitating a wide range of downstream applications. However, we argue that such formulations hinder text-to-behavior alignment. By conditioning on state-action pairs rather than actions alone, these methods require jointly modeling text-to-state and text-to-action correspondences, which introduces interference between states and actions representations. This leads to entangled optimization objectives, thereby weakening text-to-action alignment.
\section{Preliminaries}
\paragraph{Physics Simulation Setup}
We simulate a physics-based character with an SMPL-like~\cite{loper2023smpl} skeleton in the Isaac Gym~\cite{liang2018gpu}. The humanoid comprises 24 joints and 69 actuators, each driven by a proportional-derivative (PD) controller. At each time step, the policy outputs an action $\bm{a}_t \in \mathbb{R}^{69}$, which specifies the target joint positions for all actuators. Given the current state $\bm{s}_t \in \mathbb{R}^{358}$, the simulator advances the system to the next state $\bm{s}_{t+1}$ according to the dynamic transition $\bm{s}_{t+1} \sim p(\bm{s}_{t+1}\mid\bm{s}_t, \bm{a}_t)$.

\paragraph{Data Collection} 
We track reference motions from the HumanML3D dataset~\cite{guo2022generating} using the PHC tracking policy~\cite{luo2023perpetual} in a physics-based simulation environment. According to the success criteria defined in PHC~\cite{luo2023perpetual}, only successfully tracked trajectories are collected, ensuring high-quality physically plausible expert demonstrations. Each retained trajectory is then paired with its corresponding text annotation, forming state-action-text triplets. The resulting datasets are subsequently used to train a diffusion policy via behavior cloning, enabling it to learn mappings from textual commands and humanoid states to corresponding actions.

\paragraph{Humanoid State Representation}
We define the humanoid state as $\bm{s} = (h, \bm{p}, \bm{r}, \bm{v}, \bm{\omega})$, where $h\in \mathbb{R}$ denotes the height of the humanoid's root joint above the ground plane. $\bm{p} \in \mathbb{R}^{23 \times 3}$ represents the positions of all non-root joints in the humanoid root coordinate frame. $\bm{r} \in \mathbb{R}^{24\times6}$ encodes the local orientations using a continuous 6D rotation representation. Additionally, $\bm{v}\in \mathbb{R}^{24\times3}$ and $\bm{\omega}\in\mathbb{R}^{24\times3}$ correspond to the local linear and angular velocities of each joint, respectively. The root joint is defined at the humanoid's pelvis.

\begin{table*}
\caption{\textbf{Quantitative comparison of text-driven physics-based humanoid control} on the HumanML3D~\cite{guo2022generating}  test set, where $\pm$ indicates 95\% confidence interval and $\rightarrow / \uparrow / \downarrow$ means the closer / larger / smaller the better. \textbf{Bold} and \underline{underline} indicate the best and second-best results, respectively. Phys-GT refers to the physics-based trajectories that are tracked from the MoCap dataset.}
\label{table:comparison}
\centering
\resizebox{1.0\textwidth}{!}{
\begin{tabular}{lccccccc}
\toprule
&\multicolumn{3}{c}{R-precision $\uparrow$} &  \multirow{2}{*}{FID $\downarrow$} & \multirow{2}{*}{MM Dist $\downarrow$} & \multirow{2}{*}{Diversity $\rightarrow$} & \multirow{2}{*}{MModality $\uparrow$}\\
\cmidrule(lr){2-4}
& Top 1 & Top 2 & Top 3 &  \\
\midrule
Phys-GT & $0.5590^{\pm 0.0031}$ & $0.7442^{\pm 0.0027}$ & $0.8259^{\pm 0.0023}$ & $0.0001^{\pm 0.000}$ & $1.2839^{\pm 0.0005}$ & $1.2355^{\pm 0.0083}$ & -  \\
\midrule
PDP ~\cite{truong2024pdp} & $0.0324^{\pm 0.0013}$ & $0.0646^{\pm 0.0011}$ & $0.0967^{\pm 0.0025}$ & $1.6199^{\pm 0.0146}$ & $2.4715^{\pm 0.0027}$ & $0.9713^{\pm 0.0091}$ & $\mathbf{1.8902}^{\pm 0.0013}$ \\
 CLoSD~\cite{tevet2024closd}  & $0.1717^{\pm 0.0018}$ & $0.2723^{\pm 0.0017}$ & $0.3468^{\pm 0.0026}$ & $1.5416^{\pm 0.0044}$ & $2.0941^{\pm 0.0021}$ & $0.9806^{\pm 0.0074}$ & $0.3727^{\pm 0.0537}$ \\
 UniPhys~\citep{wu2025uniphys} & $0.0865^{\pm 0.0019}$ & $0.1490^{\pm 0.0020}$ & $0.2053^{\pm 0.0029}$ & $0.6028^{\pm 0.0068}$ & $2.2846^{\pm 0.0025}$ & $1.1524^{\pm 0.0064}$ & $\underline{1.7484}^{\pm 0.0010}$  \\
Kimodo++~\cite{rempe2026kimodo}  & $\underline{0.3860}^{\pm 0.0046}$ & $\underline{0.5591}^{\pm 0.0060}$ & $\underline{0.6617}^{\pm 0.0054}$ & $\underline{0.2680}^{\pm 0.0043}$ & $\underline{1.6086}^{\pm 0.0022}$ & $\underline{1.2047}^{\pm 0.0086}$ & $1.1847^{\pm 0.0025}$  \\
\midrule
MIND (Ours) & $\mathbf{0.4679}^{\pm 0.0032}$ & $\mathbf{0.6374}^{\pm 0.0054}$ & $\mathbf{0.7299}^{\pm 0.0042}$ & $\mathbf{0.1184}^{\pm 0.0009}$ & $\mathbf{1.4576}^{\pm 0.0039}$ & $\mathbf{1.2278}^{\pm 0.0073}$ & $0.9864^{\pm 0.007}$ \\
\bottomrule
\end{tabular}}
\end{table*}

\section{Method}

\subsection{Overview}
 Our goal is to control a humanoid to perform natural, physically plausible behaviors that are consistent with the given textual commands. To this end, we propose MIND (Fig.~\ref{fig:overview_pipeline}), a novel end-to-end diffusion framework for text-driven physics-based humanoid control, which integrates multi-scale intent modeling into an action diffusion transformer (ADiT). 
 To obtain compact intent representations, we first train a variational autoencoder (VAE) to encode the high-dimensional states into a latent space (\cref{sec:LatentSpace}). Building upon this latent representation, we introduce a multi-scale intent mechanism (Fig.~\ref{fig:HIIP}) that models behavioral intent at different temporal granularities (\cref{sec:MIDF}). Specifically, a holistic intent predictor (HIP) captures global and long-horizon intent, while an immediate intent predictor (IIP) provides fine-grained, step-wise intent. Both HIP and IIP are jointly trained with the action diffusion transformer within a unified framework. This formulation enables the generation of actions that are coherent, physically grounded, and aligned with the underlying intent.
 
\subsection{Intent Variational Autoencoder}\label{sec:LatentSpace}

To capture the underlying intent embedded in humanoid state to facilitate subsequent learning, in practice, we employ a 1D causal convolutional architecture~\cite{xiao2025motionstreamer} to construct a temporal encoder-decoder that maps raw state sequences into a compact latent space. The use of causal convolutions preserves temporal causality, making the learned latent representation more suitable for sequential behavior modeling. Given a state clip $S\in\mathbb{R}^{N\times d}$, the encoder $\mathcal{E}$ produces a sequence of latent vectors $I \in \mathbb{R}^{l \times d_l}$, which is then decoded by $\mathcal{D}$ to reconstruct the raw state sequence $\hat{S} = \mathcal{D}(I)$. We use the same loss function as $\sigma$-VAE~\cite{rybkin2021simple}
to optimize the intent VAE. The full loss function is defined as:
\begin{equation}
    \mathcal{L}_{VAE} = \mathcal{L}_{rec} + \lambda_{KL}\mathcal{L}_{KL}.
\end{equation}
where 
$\mathcal{L}_{rec} = || S - \hat{S} ||_2^2$ denotes the standard VAE reconstruction loss and $\mathcal{L}_{KL} = D_{KL}(q(I|S) || p(I))$ is the KL divergence. $\lambda_{KL}$ is the balancing hyperparameter.
This formulation induces a compact and temporally coherent latent space that reduces redundancy and noise inherent in the raw state representations, thereby facilitating more effective intent modeling.
Once training is completed, the VAE is fixed, and only the encoder $\mathcal{E}$ is retained to extract intent representations for the subsequent stage. 
\subsection{Multi-Scale Intent Mechanism}\label{sec:MIDF}

To achieve semantically grounded and intent-aware humanoid control, we propose a multi-scale intent mechanism that consists of two components: the HIP and IIP. Given textual commands, the HIP predicts global intent representations $\tilde{I_H}$ that serve as shared global contextual guidance for both the IIP and ADiT. Guided by the commands, the history intent, and the predicted global intent from HIP, the IIP generates step-wise intents at a finer temporal scale. Finally, the ADiT generates physically plausible and text-aligned behaviors conditioned on the resulting multi-scale intent representation.
\paragraph{Holistic Intent Predictor}

\begin{figure}
    \centering
    \includegraphics[width=\columnwidth]{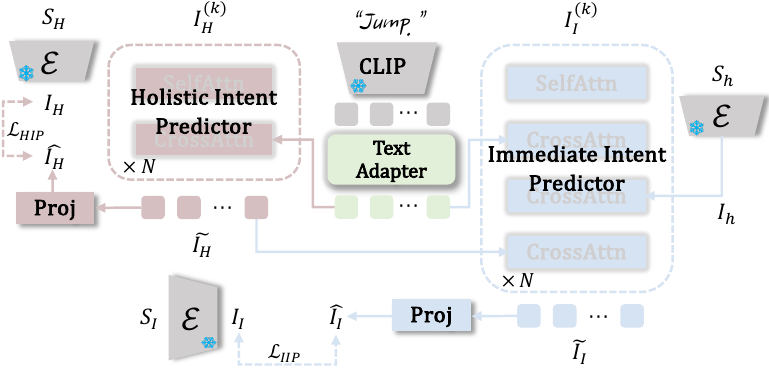}
    
    \caption{
    \textbf{Multi-scale intent mechanism.} 
 Given a textual command, a frozen CLIP text encoder with a lightweight text adapter first extracts semantic representations, which are then simultaneously injected into two complementary branches via cross-attention: a Holistic Intent Predictor for capturing global, sequence-level intent, and an Immediate Intent Predictor for modeling fine-grained, step-wise dynamics. Meanwhile, a pretrained VAE encoder is employed to encode the states into a compact latent space to capture the underlying intent. The light pink lines indicate processes related to HIP, while the light blue lines correspond to those related to IIP.
    }
    \label{fig:HIIP}
\end{figure}

As mentioned above, the HIP provides high-level holistic guidance for humanoid command execution. Given a full state sequence $S_H=(s_1, s_2, ..., s_n)\in \mathbb{R}^{n\times 358}$, we encode it into a compact latent representation using the frozen VAE encoder $\mathcal{E}$, yielding holistic intent $I_H=\mathcal{E}(S_H) \in \mathbb{R}^{l_H\times d_l}$. As illustrated in Fig.~\ref{fig:HIIP}, we train a diffusion model $\mathcal{M}_H$ to generate holistic intent conditioned on textual input. Specifically, the model predicts the denoised holistic intent sequence as $\mathcal{M}_H(I_H^k, k, c)$, where $I_H^k$ is a holistic intent sequence with added Gaussian noise, $k$ is the diffusion step, and $c$ represents the text conditioning tokens. The training loss is given by:
\begin{equation}
    \mathcal{L}_{HIP} = \mathbb{E}_{k, I_H^0}[|| I_H^0 - \mathcal{M}_H(I_H^k, k, c) ||^2],
\end{equation}
where $I_H^0 = I_H $ is the clean holistic intent sequence.
\paragraph{Immediate Intent Predictor}

Relying solely on coarse-grained holistic intent to guide action generation may compromise fine-grained control over the underlying dynamics. As the global holistic intent remains temporally invariant, it does not specify which components are relevant at each timestep, resulting in ambiguous credit assignment and blurry supervision. To address this limitation, we introduce an immediate intent predictor that provides step-wise, fine-grained guidance. As illustrated in Fig.~\ref{fig:HIIP}, at timestep $t$, we define a history state sequence $S_h = (s_{t-h}, ..., s_t)$ and a short-horizon future state sequence $S_I = (s_{t+1}, ..., s_{t+L})$. 
The corresponding immediate intent is obtained as $I_I = \mathcal{E}(S_I)$. 
We encode $S_h$ into $I_h = \mathcal{E}(S_h)$, and train an autoregressive diffusion model $\mathcal{M}_I$ to predict $I_I$ conditioned on text tokens $c$, $I_h$, and $\tilde{I_H}$. The training loss of IIP is defined as: 
\begin{equation}
    \mathcal{L}_{IIP} = \mathbb{E}_{k, I_I^0}[|| I_I^0 - \mathcal{M}_I(I_I^k, k, c, I_h, \tilde{I_H}) ||^2],
\end{equation}
where $I_I^k$ is an immediate intent sequence clip with added Gaussian noise, $k$ is the diffusion step,  $I_I^0 = I_I $ is the clean immediate intent sequence clip. In practice, $\tilde{I_H}$ is instantiated as the final-layer hidden representation of the HIP.
\begin{table}
\caption{\textbf{Ablation study on the effect of each design choice.}}
\label{table:ablation}
\centering
\resizebox{1.0\columnwidth}{!}{
\begin{tabular}{ccccccccc}
\toprule
\multirow{2}{*}{AdaLN}&\multirow{2}{*}{CrossAttn}&\multirow{2}{*}{IIP}&\multirow{2}{*}{HIP}&\multirow{2}{*}{VAE}&\multicolumn{3}{c}{R-precision $\uparrow$} &  \multirow{2}{*}{FID $\downarrow$} \\

 & & & & & Top 1 & Top 2 & Top 3 &  \\
\midrule

\ding{51} & & & & & 0.1735 &0.2827 & 0.3608 & 0.7513 \\

& \ding{51}& & & & 0.3155 & 0.4645 & 0.5530  & 0.3632\\

& \ding{51}&\ding{51} & & &  0.3233&0.4813 &0.5759  & 0.3314 \\

& \ding{51}&\ding{51} & &\ding{51} &$\underline{0.3601}$  &$\underline{0.5174}$ &$\underline{0.6084}$  &$\underline{0.2007}$ \\

& \ding{51}& &\ding{51} &\ding{51} &  0.2917&0.4372 &0.5273  & 0.5362 \\

& \ding{51}&\ding{51} &\ding{51} &\ding{51} & $\mathbf{0.4679}$  &$\mathbf{0.6374}$ &$\mathbf{0.7299}$  &$\mathbf{0.1184}$\\
\bottomrule
\end{tabular}}
\end{table}

\subsection{Multi-Scale Intent Informed Diffusion Policy}

Based on the HIP and IIP, we leverage the multi-scale intent to jointly inform action generation. The holistic intent captures long-horizon planning, while the immediate intent provides step-wise refinement. This hierarchical formulation enables a comprehensive modeling of humanoid behavioral dynamics, facilitating more precise and temporally consistent control. As shown in Fig.~\ref{fig:overview_pipeline}, ADiT adopts the diffusion transformer (DiT)~\cite{peebles2023scalable} as its backbone. Text tokens, along with the holistic intent and immediate intent representations, are incorporated via cross attention to inform action synthesis. Formally, given the history state $S_h$, we train an autoregressive DiT $\mathcal{M}_A$ to generate actions conditioned on text and multi-scale intent. The training loss of ADiT is formulated as: 
\begin{equation}
    \mathcal{L}_{ADiT} = \mathbb{E}_{k, a^0}[|| a^0 - \mathcal{M}_A(a^k, k, c, S_h, \tilde{I_H},\tilde{I_I}) ||^2],
\end{equation}
where $a^k$ denotes the noisy action at diffusion step $k$, $ a^0$ is the clean action sequence, $\tilde{I_H}$ and $\tilde{I_I}$ denote the final-layer hidden representations from the HIP and the IIP, respectively. In contrast to intent modeling, the ADiT directly operates on state history sequence $S_h$ without latent projection. This design choice preserves fidelity of humanoid states, enabling accurate perception of the current behavioral dynamics essential for physically grounded control.

\paragraph{Training and Inference} 
During training, all components, including the HIP, IIP, and ADiT, are jointly optimized in an end-to-end manner, as illustrated in Fig.~\ref{fig:overview_pipeline}:
\begin{equation}
    \mathcal{L} = \mathcal{L}_{HIP} + \mathcal{L}_{IIP} + \mathcal{L}_{ADiT}.
\end{equation}
This unified training paradigm promotes consistent representation learning across intent and action spaces, facilitates information flow across temporal scales, and enables mutual refinement between high-level intent planning and low-level motor action generation.

For inference, as shown in Fig.~\ref{fig:overview_pipeline}, the model continuously receives states from the simulator and stores them in a first-in-first-out history buffer. Given the textual command, the HIP generates the corresponding holistic intent to provide global behavioral guidance. In parallel, the IIP first encodes the history states stored in the buffer into a compact latent space, and then predicts the corresponding immediate intent conditioned on both the history latent representations and the textual tokens. ADiT jointly takes the predicted holistic intent, immediate intent and the given command as conditions to generate the corresponding actions. The predicted actions are then executed to drive the humanoid, repeated autoregressively for the desired steps.

\section{Experiments}

\begin{figure}
    \centering
    \includegraphics[width=\columnwidth]{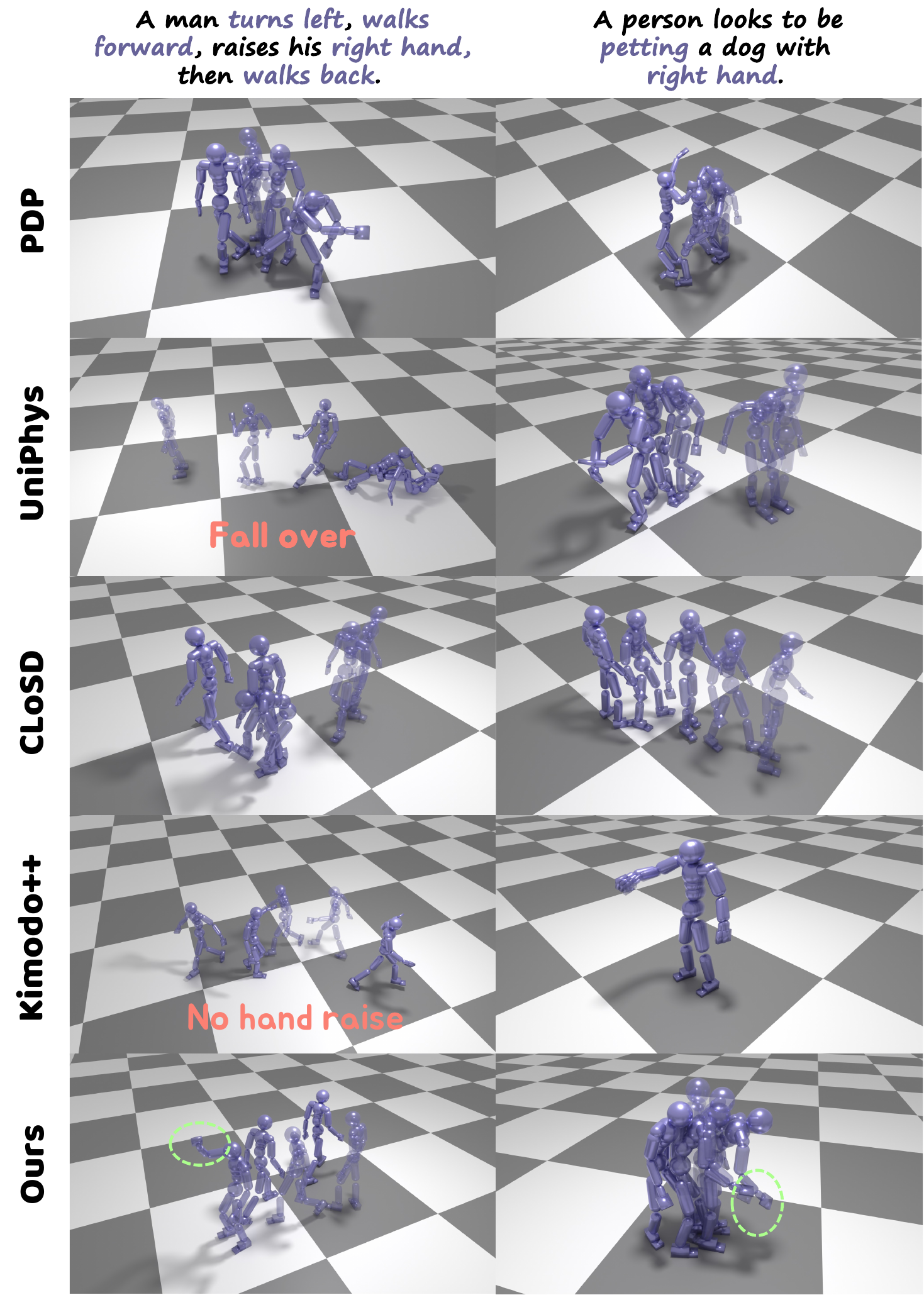}
    
    \caption{
    \textbf{Qualitative comparison.} Our MIND generates coherent and natural humanoid behaviors with stronger alignment to textual inputs, outperforming all existing baselines.
    }
    \label{fig:qualitative comparison}
\end{figure}

\subsection{Experiment Setup}
 \begin{table}
\caption{
\textbf{Quantitative evaluation of physical correctness.} Following~\cite{wu2025uniphys,tevet2024closd,yuan2023physdiff}, we report foot-floating to assess the physical plausibility and motion jerk to evaluate the smoothness.
}
\label{table:physics_metric_comparsion}
\centering
\resizebox{1.0\columnwidth}{!}{
\begin{tabular}{lcc}
\toprule
 & Floating $[\mathrm{mm}]$ $\downarrow$ & Jerk $[\mathrm{mm/frame^3}]$ $\downarrow$ \\
\midrule
Phys-GT & 15.5840 & $2.7172\times10^{-3}$ \\
\midrule
PDP~\cite{truong2024pdp} &24.8511 & $6.5476\times10^{-3}$ \\
CLoSD~\cite{tevet2024closd} & \underline{19.2958} &  $12.0022\times10^{-3}$ \\
UniPhys~\cite{wu2025uniphys} & 20.4853 &  $\underline{3.6054\times10^{-3}}$ \\
Kimodo++~\cite{rempe2026kimodo} &33.7176 & $10.2709\times 10^{-3}$   \\
\midrule
MIND (Ours) & $\mathbf{17.1173}$ &  $\mathbf{2.5966\times10^{-3}}$ \\
\bottomrule
\end{tabular}}
\end{table}
\paragraph{Dataset and evaluation metrics}
We evaluate our model on the HumanML3D~\cite{guo2022generating} dataset, a large-scale human motion capture dataset with fine-grained textual annotations, where the vast majority of motions are sourced from the AMASS dataset~\cite{mahmood2019amass}. We follow the standard evaluation protocol commonly used in text-to-motion literature, which includes five metrics: (1) \textit{R-Precision} evaluates text-motion consistency by measuring motion-to-text retrieval accuracy. (2) \textit{Fréchet Inception Distance (FID)} measures the distributional discrepancy between generated motions and ground truth. (3) \textit{Multimodal Distance (MM Dist)} assesses the alignment between text and its corresponding motion in the shared latent space. (4) \textit{Diversity} quantifies the variability across different generated sequences. (5) \textit{Multimodality (MModality)} evaluates the variation among motions generated from the same textual input. To compute these metrics, we train a behavior encoder and a text encoder using a contrastive learning objective, which maps paired text-state samples to a shared latent space. 
Following PhysDiff~\cite{yuan2023physdiff}, CLoSD~\cite{tevet2024closd} and UniPhys~\cite{wu2025uniphys}, we further evaluate the physical plausibility of the generated motions using foot-floating and jerk, which measure stability and smoothness, respectively. We do not report foot skating or penetration, as they are negligible across all physics-based methods. For a fair comparison, all methods (except Kimodo++) are initialized from a standardized neutral standing pose during evaluation.

\begin{figure}
    \centering
    \includegraphics[width=\columnwidth]{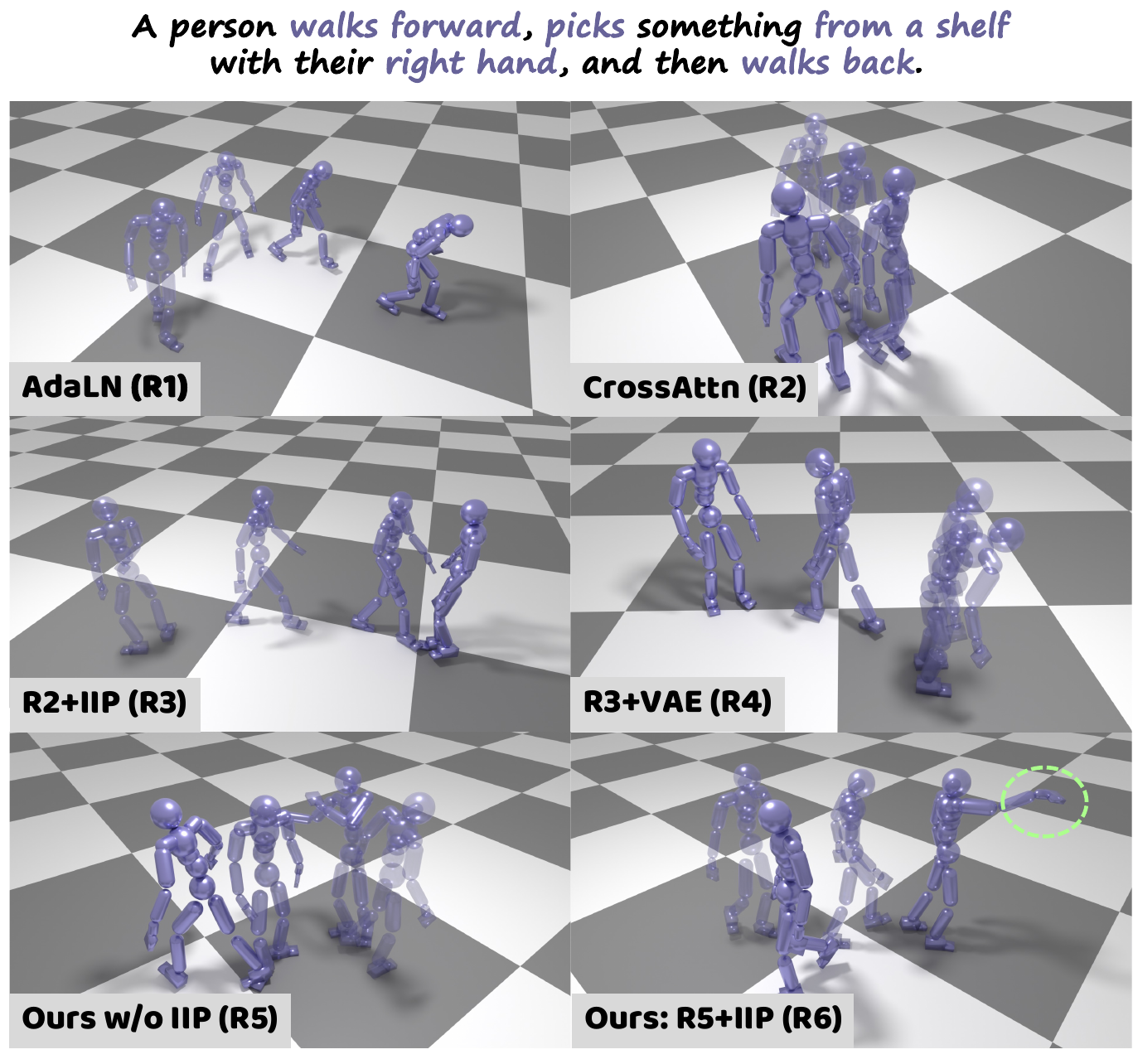}
    
    \caption{
    \textbf{Qualitative ablation study.} We qualitatively evaluate the effect of each component of our proposed MIND. In each subfigure, bold black text in parentheses indicates the corresponding row index in Tab.\ref{table:ablation}
    }
    \label{fig:qualitative abltion}
\end{figure}

\subsection{Comparison}
We compare our method against four baselines. PDP~\cite{truong2024pdp} and UniPhys~\cite{wu2025uniphys} directly generate actions conditioned solely on textual inputs in an end-to-end manner. CLoSD~\cite{tevet2024closd} employs a diffusion-based planner followed by an RL-based tracking controller to execute the planned kinematic motions. Kimodo++ first synthesizes kinematic trajectories from textual commands using Kimodo~\cite{rempe2026kimodo}, and then employs the PHC+~\cite{luo2023universal} tracking policy, which achieves nearly $100\%$ success rate on AMASS~\cite{mahmood2019amass}, to obtain physically plausible motions.

\begin{figure}
    \centering
    \includegraphics[width=\columnwidth]{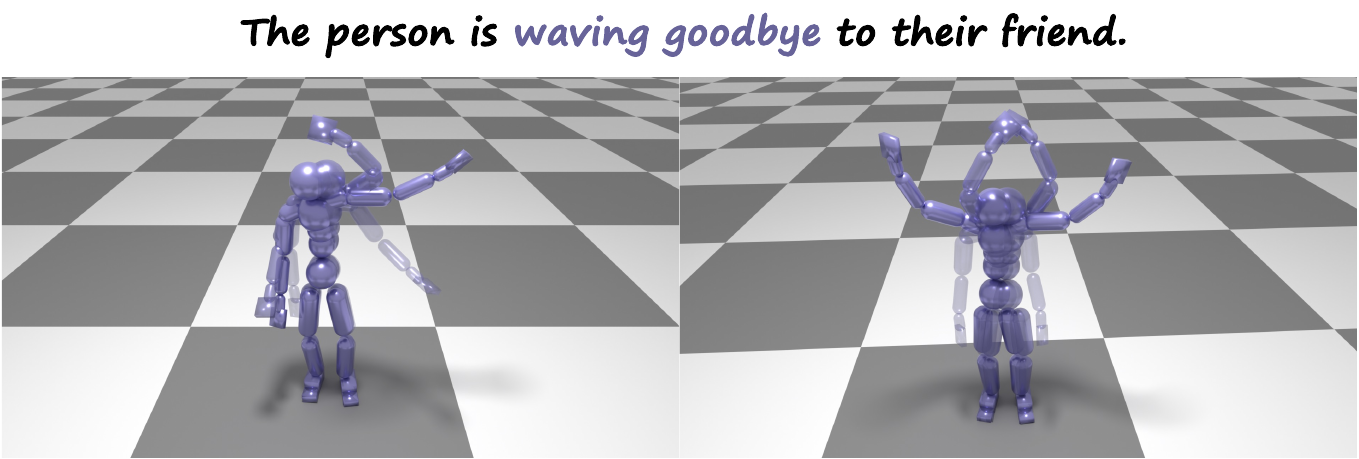}
    
    \caption{
    \textbf{Behavioral diversity under the same textual command.} Given an identical text prompt, MIND is able to generate diverse humanoid behaviors that are all semantically consistent while exhibiting different motion patterns and dynamics.
    }
    \label{fig:diversity}
\end{figure}

\paragraph{Quantitative Results} As shown in Tab.~\ref{table:comparison}, our proposed MIND achieves consistently strong performance across most evaluation metrics, demonstrating a favorable balance between semantic alignment, behavior quality, and diversity. Notably, while MIND lags behind PDP~\cite{truong2024pdp} and UniPhys~\cite{wu2025uniphys} on the MModality metric, this is primarily because these end-to-end methods suffer from weak text conditioning, leading to nearly random generation, which significantly increases variability without meaningful semantic control. 
 For physical evaluation metrics, shown in Tab.~\ref{table:physics_metric_comparsion}, MIND outperforms all baseline methods, achieving the lowest floating error and even surpassing Phys-GT in jerk. Notably, even though Kimodo~\cite{rempe2026kimodo} is trained on large-scale datasets and coupled with the strong RL-based tracker, its performance remains suboptimal. We also observe that methods based on the kinematic–tracking paradigm consistently exhibit higher jerk score than the end-to-end methods. This may be attributed to the domain shift introduced by the kinematic-tracking paradigm, where kinematic generation and tracking are decoupled and optimized independently, as also noted in prior works~\cite{huang2025diffuse, wu2025uniphys}. This evidence further substantiates the superiority of our framework, which jointly models intent and action in a unified formulation, effectively eliminating such mismatch.
\paragraph{Qualitative Results}
As shown in Fig.~\ref{fig:qualitative comparison}, PDP~\cite{truong2024pdp}, UniPhys~\cite{wu2025uniphys}, and CLoSD~\cite{tevet2024closd} exhibit poor text-behavior alignment, with the generated behaviors often failing to correspond to the input textual commands. Kimodo++, benefiting from the powerful kinematic motion generation capability of Kimodo~\cite{rempe2026kimodo}, can only roughly produce the expected actions, yet still suffers from noticeably unnatural behaviors and incomplete semantic execution. For example, in the first generated case of Kimodo++, the model fails to perform the “raise right hand” behavior, while in the second case, the generated “petting” behavior appears particularly rigid and unnatural. In contrast, our method is able to generate highly vivid, semantically aligned and physically consistent humanoid behaviors, demonstrating that our multi-scale intent framework enables effective and robust guidance for humanoid command execution. Moreover, as shown in Fig.~\ref{fig:diversity}, given identical textual commands, our proposed MIND is capable of generating diverse behavior patterns. For instance, for the concept of "waving goodbye" , both one-hand waving and two-hand waving are valid behavioral realizations, indicating that our method not only achieves strong semantic alignment but also captures rich and semantically grounded behavioral multimodality.
 \begin{table}
\caption{\textbf{Ablation study on physics-based metrics.}}
\label{table:physics_metric_ablation}
\centering
\resizebox{1.0\columnwidth}{!}{
\begin{tabular}{ccccccc}
\toprule
{AdaLN}&{CrossAttn}&{IIP}&{HIP}&{VAE}
 & Floating $[\mathrm{mm}]$ $\downarrow$ & Jerk $[\mathrm{mm/frame^3}]$ $\downarrow$\\
\midrule

\ding{51} & & & & & 20.8682 & $3.8597\times10^{-3}$ \\

& \ding{51}& & & & 20.1482 & $3.3839\times10^{-3}$ \\

& \ding{51}&\ding{51} & & &  19.3830&$2.8497\times10^{-3}$ \\

& \ding{51}&\ding{51} & &\ding{51} &$\underline{17.4492}$  &$\underline{2.6982\times10^{-3}}$  \\

& \ding{51}& &\ding{51} &\ding{51} &  21.5535&$4.3387\times10^{-3}$  \\

& \ding{51}&\ding{51} &\ding{51} &\ding{51} & $\mathbf{17.1173}$  &$\mathbf{2.5966\times10^{-3}}$\\
\bottomrule
\end{tabular}}
\end{table}

\subsection{Ablation Study}

\paragraph{Effect of Text Condition Mechanism}
We first investigate the impact of two text conditioning mechanisms, global token adaptive layer normalization~\cite{peebles2023scalable} and multi-token cross attention, as shown in the first two rows of Tab.~\ref{table:ablation}. It is observed that multi-token cross attention significantly outperforms the single global token adaptive layer normalization. This improvement likely stems from the richer and more expressive semantic representations provided by multiple tokens, which enable more precise conditioning for action generation. This also suggests that the suboptimal text-semantic alignment observed in PDP~\cite{truong2024pdp} and UniPhys~\cite{wu2025uniphys}, which share the same end-to-end paradigm as our proposed MIND, may be partly attributed to their reliance on single global token adaptive layer normalization for text conditioning, which limits the expressiveness of textual representations.
\paragraph{Effect of Multi-scale Intent}
As shown in Tab.~\ref{table:ablation} and Fig.~\ref{fig:qualitative abltion}, row 1 (R1) and R3 demonstrate that the immediate intent predicted by the IIP effectively provides fine-grained guidance for action generation. R3 and R4 indicate that a more compact latent space facilitates better semantic intent modeling, leading to further performance gains. R4, R5 and R6 verify that our proposed multi-scale intent mechanism effectively integrates long-term planning with immediate expectations, substantially improving overall performance. Notably, comparing R2 and R5, it is observed that relying solely on holistic intent to guide action generation instead degrades performance. This may stem from its tendency to overemphasize global average structure while lacking sufficient temporal specificity, thereby undermining the fine-grained dynamics required at each timestep. Furthermore, as shown in Tab.~\ref{table:physics_metric_ablation}, each part of the proposed multi-scale intent mechanism also consistently improves all physics-based metrics. In particular, the comparison between R2 and R3 demonstrates that incorporating the immediate intent contributes to alleviate both floating artifacts and motion jerkiness, highlighting the effectiveness of the IIP in refining motion dynamics. Meanwhile, R3 and R4 further suggest that modeling intent in a compact latent space can mitigate the negative impact of redundancy and noise inherent in raw state representations, leading to more stable and physically plausible behaviors.

\begin{figure}
    \centering
    \includegraphics[width=\columnwidth]{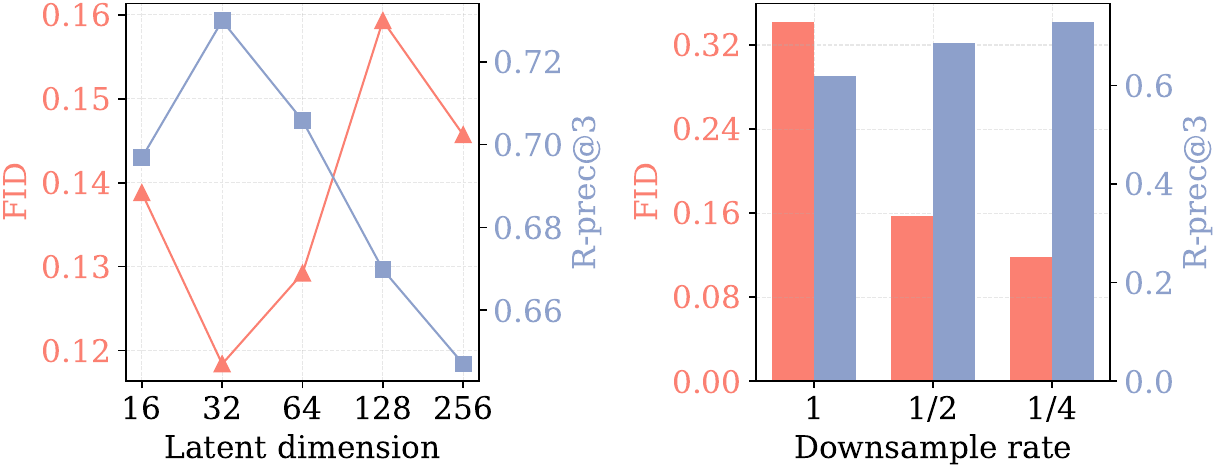}
    
    \caption{
    \textbf{Effect of latent dimension and temporal downsampling rate.} 
    }
    \label{fig:ablation vae}
\end{figure}

\paragraph{Effect of Latent Size}
We conduct an exploration on both latent dimension and downsampling rate. As illustrated in the left panel of Fig.~\ref{fig:ablation vae}, a latent dimension of 32 achieves the best performance in terms of both R-Precision and FID. Increasing the dimension leads to slight degradation, while reducing it to 16 results in insufficient representational capacity and thus inferior performance. In the right panel, with the latent dimension fixed at 32, it is observed that a $1/4$ downsampling rate significantly outperforms both $1/2$ downsampling and no downsampling. Overall, these results suggest that a compact latent space facilitates improving semantic alignment between behavioral intent and textual commands, thereby enabling more semantically consistent action generation.

\section{Conclusions}
In this work, we presented MIND, a novel end-to-end diffusion framework for text-driven physics-based humanoid control. Unlike existing approaches that either suffer from the domain discrepancy between kinematic motion generation and physics-based tracking or struggle with the substantial modality gap between textual commands and low-level actions, our proposed MIND explicitly models behavioral intent as an intermediate semantic representation to bridge text and action under a unified diffusion framework. By introducing a multi-scale intent mechanism that captures humanoid intent at different temporal granularities, MIND combines holistic behavioral guidance with step-wise intent refinement for more semantically aligned and physically plausible behavior synthesis. Furthermore, by encoding humanoid states into a compact latent space, MIND obtains more structured intent representations, further enhancing semantic expressiveness and text-action alignment. Comprehensive quantitative and qualitative analysis demonstrates that our proposed MIND consistently outperforms all existing baselines in semantic alignment and physical realism. We believe this work can inspire future research on physics-based humanoid control and facilitate the development of more natural, controllable, and semantically aware physics-based character animation systems.

\section{Acknowledgement}
This project was supported by Shanghai Local College Capacity Building Program (23010503100), NSFC (No.62406195, W2431046, W2431046), Shanghai Frontiers Science Center of Human-centered Artificial Intelligence (ShangHAI), MoE Key Laboratory of Intelligent Perception and Human-Machine Collaboration (ShanghaiTech University), the Shanghai Frontiers Science Center of Human-centered Artificial Intelligence, HPC Platform and Core Facility Platform of Computer Science and Communication of ShanghaiTech University and Shanghai Engineering Research Center of Intelligent Vision and Imaging.

\bibliographystyle{ACM-Reference-Format}
\bibliography{main}

\clearpage
\appendix
\section{Additional Implementation Details}
In practice, the HIP, IIP, and ADiT are implemented with 6 DiT~\cite{peebles2023scalable} blocks whose latent dimension is set to 768, and each attention layer consists of 8 heads. For the Intent VAE, both the encoder and decoder are based on 1D causal ResNet blocks~\cite{he2016deep}, and the temporal downsampling rate and the latent dimension are set to 4 and 32, respectively. We set $\lambda_{KL} = 1 \times 10^{-5}$.
 For HIP, each full state sequence is uniformly downsampled to 16 frames before encoding into the latent space for computational efficiency. The horizon of immediate intent is set to $L=16$, while the action prediction horizon is set to 4. The history state horizon is $h=16$. To ensure consistent intent representations, HIP and IIP are formulated within a shared latent space.
 We use CLIP-ViT-L/14~\cite{radford2021learning} as the text encoder, and a two-layer transformer encoder as the text adapter. We further adopt classifier-free guidance~\cite{ho2022classifier} by randomly masking 10\% of text tokens during training, and apply a guidance scale of 3.5 during diffusion sampling. All our models are trained for 80K iterations with batch size of 448 on 8 NVIDIA GeForce RTX 4090 GPUs, requiring approximately 12 hours in total. 

\section{Failure Analysis of Existing End-to-End Baselines}
Although two-stage approaches generally suffer from performance degradation due to the domain shift issue discussed in the main paper, an important question remains: why PDP~\cite{truong2024pdp} and UniPhys~\cite{wu2025uniphys}, which both follow an end-to-end paradigm, exhibit particularly poor performance? As shown in the quantitative comparison Tab.1 of the main paper, the R-Precision of PDP is nearly indistinguishable from random generation (the evaluation batch size is 32). We further study the reasons behind this result. Specifically, we carefully examine the official implementation of PDP, which introduces textual conditioning through FiLM modulation~\cite{perez2018film}. However, we find that the textual conditioning is applied exclusively to the history state representation, without any explicit fusion of textual semantics into the action. As a result, the model is unable to effectively establish text-to-action controllability. For UniPhys, textual information is incorporated via adaptive layer norm~\cite{peebles2023scalable}. However, it employs a shared transformer decoder to jointly model states and latent actions, which weakens the direct effect of textual semantics on actions and instead biases the model toward learning the mapping between text and the joint state-action representation. Since states and actions inherently exhibit a modality gap, their joint modeling inevitably introduces mutual interference, which further hinders fine-grained text-action alignment.

\section{Discussion}

\paragraph{Limitations and failure cases}

\begin{figure}
    \centering
    \includegraphics[width=\columnwidth]{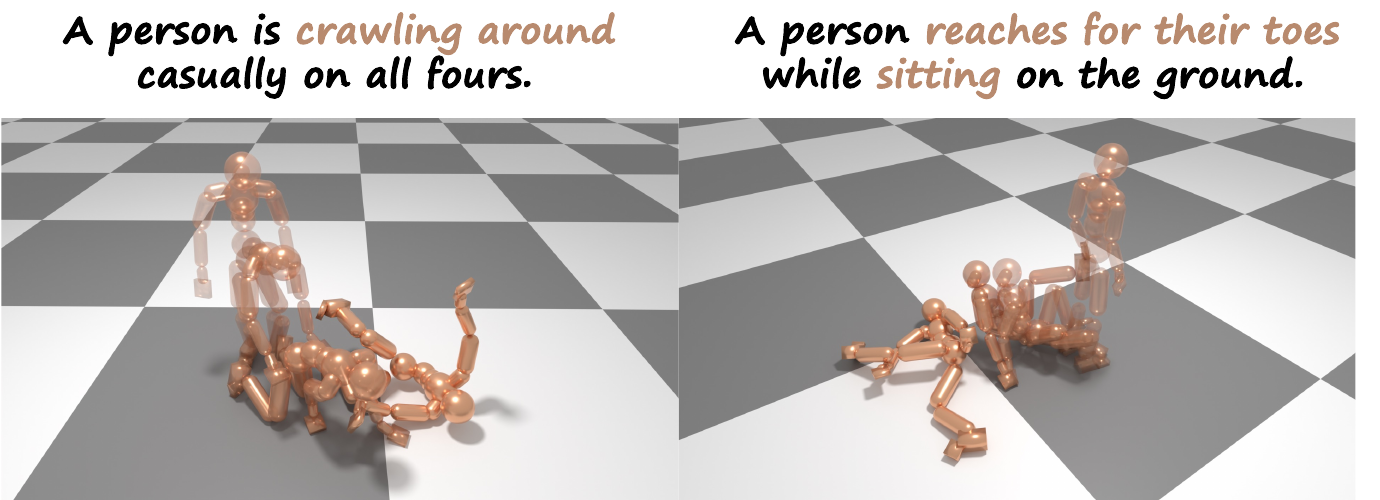}
    
    \caption{
    \textbf{Failure cases.} Our method shows limitations in highly dynamic behaviors such as crawling.
    }
    \label{fig:failure case}
\end{figure}

Although the proposed framework achieves strong performance across most cases, several highly challenging behaviors remain difficult to execute robustly, as illustrated in Fig.~\ref{fig:failure case}. In particular, behaviors that require simultaneous multi-limb coordination and precise balance control still present significant challenges for our method. For example, in the "crawling around" case on the left side of Fig.~\ref{fig:failure case}, our approach is initially able to generate behaviors that closely match the textual command. However, the humanoid gradually loses balance during crawling and eventually falls. Similarly, in the "reaches for their toes" case, the humanoid is required to sit on the ground while supporting the body with both hands and lifting one leg simultaneously, which introduces extremely unstable body configurations and often leads to failure. These failure cases suggest that a unified policy may still struggle to fully capture highly specialized balance strategies required by certain complex behaviors. One possible direction is to introduce expert-specialized modules, similar to a mixture-of-experts (MoE) framework~\cite{zhou2022mixture,mu2025comprehensive}, where different experts focus on distinct categories of challenging motions such as crawling, balancing, or acrobatic behaviors. Another promising direction is to incorporate explicit contact-aware planning, balance constraints, or differentiable stability objectives into the policy learning process to improve robustness under complex dynamics. Furthermore, integrating a predictive world model~\cite{li2025enhancing} or a dedicated recovery mechanism~\cite{luo2023perpetual} may help the humanoid better handle instability in complex dynamic behaviors.
\paragraph{Discussion and future work}

Although MIND demonstrates strong capability in generating diverse, natural, physically plausible, and text-aligned humanoid behaviors, several future directions remain worth exploring. First, the current framework mainly focuses on short-to-medium horizon behavior generation, while modeling long-horizon behaviors with complex temporal dependencies remains challenging. Incorporating hierarchical memory mechanisms or world models may further improve long-term planning and behavioral consistency. Second, although MIND achieves promising semantic alignment between textual commands and generated behaviors, handling more compositional and fine-grained language instructions is still non-trivial. Future work could explore stronger large language model (LLM)-based semantic representations or multimodal reasoning mechanisms to improve semantic understanding and controllability.
Furthermore, the diffusion model adopted in our framework inherently requires iterative denoising during inference, which limits runtime efficiency and increases computational overhead. Although our framework enables strong generative capability and stable behavior synthesis, it may hinder real-time deployment in latency-sensitive applications such as interactive control and robotics. Future work could explore model distillation and diffusion acceleration techniques to reduce inference cost while maintaining motion quality, semantic alignment, and physical realism.

\begin{figure*}[t]
    \centering
    \includegraphics[width=\textwidth]{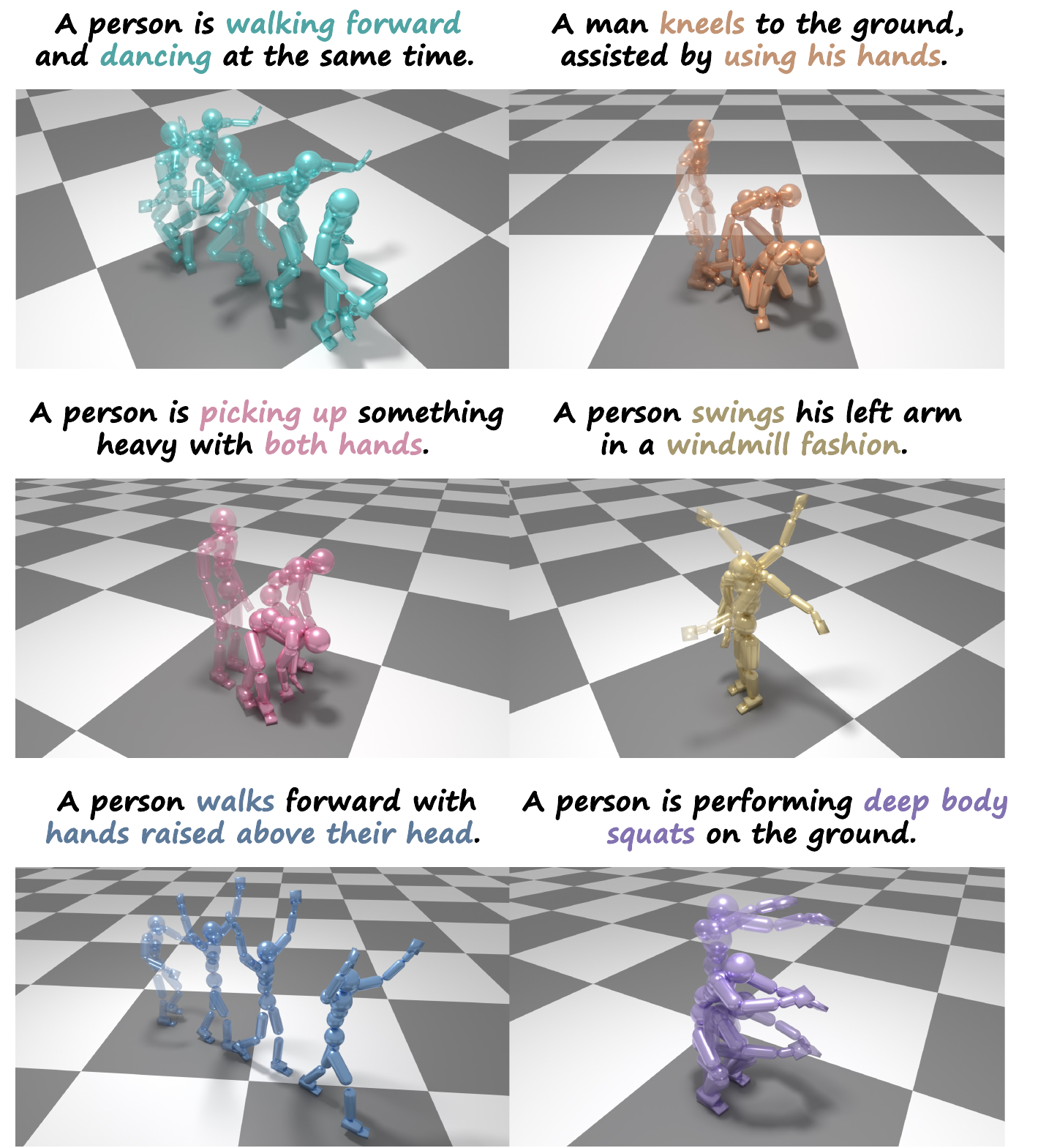}
    
    \caption{
    \textbf{Visual snapshots of text-driven humanoid behaviors via MIND.} (Part I).
    }
    \label{fig:demo_figure_part1}
\end{figure*}

\begin{figure*}[t]
    \centering
    \includegraphics[width=\textwidth]{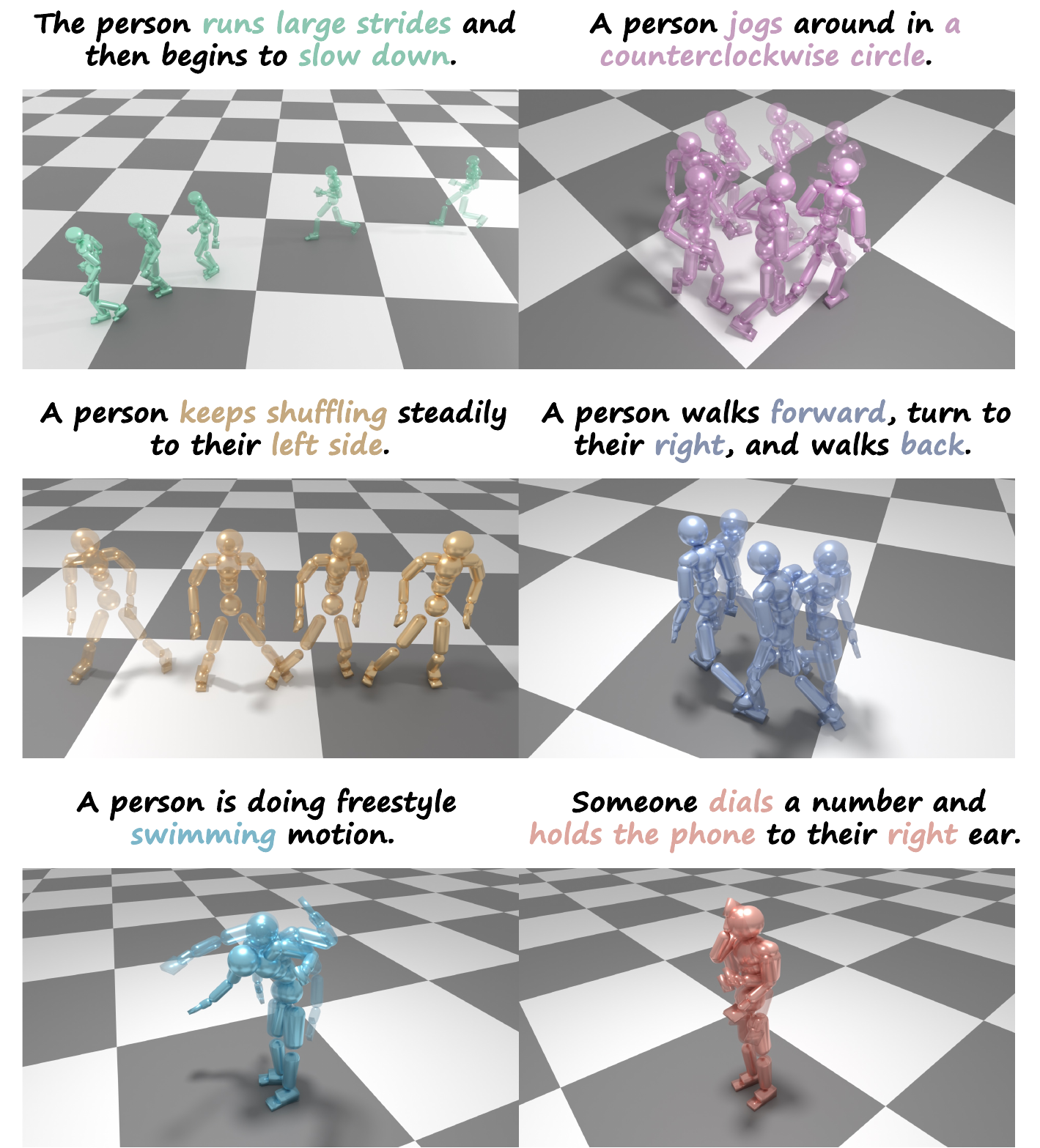}
    
    \caption{
    \textbf{Visual snapshots of text-driven humanoid behaviors via MIND.} (Part II).
    }
    \label{fig:demo_figure_part2}
\end{figure*}

\end{document}